\documentclass[final, 10pt]{IEEEtran}
\IEEEoverridecommandlockouts
\usepackage{cite}
\usepackage{amsmath,amssymb,amsfonts}
\usepackage{algorithmic}
\usepackage{graphicx}
\usepackage{textcomp}
\usepackage{xcolor}
\usepackage{float}
\def\BibTeX{{\rm B\kern-.05em{\sc i\kern-.025em b}\kern-.08em
    T\kern-.1667em\lower.7ex\hbox{E}\kern-.125emX}}

\floatstyle{plain}
\newfloat{copyrightbox}{!hb}{loc}
\floatname{copyrightbox}{Copyright Notice}

\newcommand\copyrighttext{%
\footnotesize 
\textit{This is a preprint accepted at the 6th International Conference on Artificial Intelligence and Knowledge Engineering (AIKE), 2023. \textcopyright IEEE.}
}
\begin{document}

\title{Towards Interpretable Solar Flare Prediction with Attention-based Deep Neural Networks}

\author{\IEEEauthorblockN{Chetraj Pandey, Anli Ji, Rafal A. Angryk, Berkay Aydin}\\
\IEEEauthorblockA{\textit{Department of Computer Science, Georgia State University, Atlanta, GA, USA} \\
\textit{\{cpandey1, aji1, rangryk, baydin2\}@gsu.edu}\\
} }

\maketitle
\begin{abstract}
Solar flare prediction is a central problem in space weather forecasting and recent developments in machine learning and deep learning accelerated the adoption of complex models for data-driven solar flare forecasting. In this work, we developed an attention-based deep learning model as an improvement over the standard convolutional neural network (CNN) pipeline to perform full-disk binary flare predictions for the occurrence of $\geq$M1.0-class flares within the next 24 hours. For this task, we collected compressed images created from full-disk line-of-sight (LoS) magnetograms. We used data-augmented oversampling to address the class imbalance issue and used true skill statistic (TSS) and Heidke skill score (HSS) as the evaluation metrics. Furthermore, we interpreted our model by overlaying attention maps on input magnetograms and visualized the important regions focused on by the model that led to the eventual decision. The significant findings of this study are: (i) We successfully implemented an attention-based full-disk flare predictor ready for operational forecasting where the candidate model achieves an average TSS=0.54$\pm$0.03 and HSS=0.37$\pm$0.07. (ii) we demonstrated that our full-disk model can learn conspicuous features corresponding to active regions from full-disk magnetogram images, and (iii) our experimental evaluation suggests that our model can predict near-limb flares with adept skill and the predictions are based on relevant active regions (ARs) or AR characteristics from full-disk magnetograms.
\end{abstract}

\begin{copyrightbox}
\centering
  \fbox{\parbox{\dimexpr\linewidth-5\fboxsep-3\fboxrule\relax}{\copyrighttext}}
\end{copyrightbox}
\vspace{-20pt}
\section{Introduction}
Solar flares are relatively short-lasting events, manifested as the sudden release of huge amounts of energy with significant increases in extreme ultraviolet (EUV) and X-ray fluxes, and are one of the central phenomena in space weather forecasting. They are detected by the X-ray Sensors (XRS) instrument onboard Geostationary Operational Environmental Satellite (GOES) \cite{Chamberlin2009} and classified according to their peak X-ray flux level, measured in watts per square meter ($Wm^{-2}$) into the following five categories by the National Oceanic and Atmospheric Administration (NOAA):  X $(\geq10^{-4}Wm^{-2})$, M $(\geq10^{-5}$ and $< 10^{-4}Wm^{-2})$, C $(\geq10^{-6}$ and $<10^{-5}Wm^{-2})$, B $(\geq10^{-7}$ and $<10^{-6}Wm^{-2})$, and A $(\geq10^{-8}$ and $<10^{-7}Wm^{-2})$ \cite{spaceweather}. In solar flare forecasting, M- and X-class flares are large and relatively scarce events and are usually considered to be the class of interest as they are more likely to have a near-Earth impact that can affect both space-based systems (e.g., satellite communication systems) and ground-based infrastructures (e.g.,  electricity supply chain and airline industry) and even pose radiation hazards to astronauts in space. Therefore, it is essential to have a precise and reliable approach for predicting solar flares to mitigate the associated life risks and infrastructural damages.
\begin{figure}[b!]
\vspace{-10pt}
  \centering
  \includegraphics[width=0.72\linewidth]{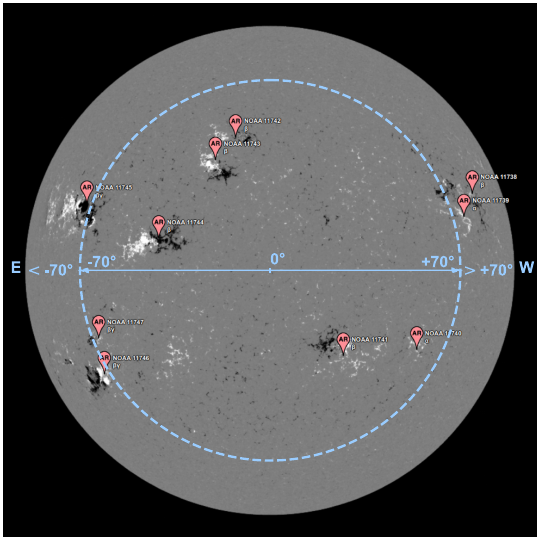}
  \caption{An annotated full-disk magnetogram image as observed on 2013-05-13 at 02:00:00 UTC, showing the approximate central location (within $\pm$70$^{\circ}$) and near-limb (beyond $\pm$70$^{\circ}$ to $\pm$90$^{\circ}$) region with all the visible active regions present at the noted timestamp, indicated by the red flags. Note that the directions East (E) and West (W) are reversed in solar coordinates.}
  \label{fig:fd_mag}
\end{figure}

Active regions (ARs) are the areas on the Sun (visually indicated by scattered red flags in full-disk magnetogram image, shown in Fig.~\ref{fig:fd_mag}) with disturbed magnetic field and are considered to be the initiators of various solar activities such as coronal mass ejections (CMEs), solar energetic particle (SEP) events, and solar flares \cite{Hamidi2019}. The majority of the approaches for flare prediction primarily target these ARs as regions of interest and generate predictions for each AR individually. The magnetic field measurements, which are the dominant feature employed by the AR-based forecasting techniques, are susceptible to severe projection effects as ARs get closer to limbs to the degree that after $\pm$60$^{\circ}$ the magnetic field readings are distorted \cite{Falconer2016}. Therefore, the aggregated flare occurrence probability (for the whole disk), in fact, is restricted by the capabilities of AR-based models. This is because the input data is restricted to ARs located in an area within $\pm$30$^{\circ}$  (e.g., \cite{Huang2018}) to $\pm$70$^{\circ}$ (e.g., \cite{Ji2020}) from the center due to severe projection effects \cite{Hoeksema2014}. As AR-based models include data up to $\pm$70$^{\circ}$, in the context of this paper, this upper limit ($\pm$70$^{\circ}$) is used as a boundary for central location (within $\pm$70$^{\circ}$)  and near-limb regions (beyond $\pm$70$^{\circ}$) as shown in Fig.~\ref{fig:fd_mag}.

Furthermore, to issue a full-disk forecast using an AR-based model, the usual approach involves aggregating the flare probabilities from each AR by applying a heuristic function, as outlined in \cite{Pandey2022f}. This aggregated result estimates the probability of at least one AR experiencing a flare event, assuming that the occurrence of flares in different ARs is conditionally independent and assigning equal weights to each AR during full-disk aggregation. This uniform weighting approach may not accurately capture the true impact of each AR on the probability of predicting full-disk flares \cite{pandey2022exploring}. It is essential to note that the specific weights for these ARs are generally unknown, and there are no established methods for precisely determining these weights. While AR-based models are limited to central locations and require a heuristic to aggregate and issue comprehensive forecasts, full-disk models use complete, often compressed, magnetograms corresponding to the entire solar disk. These magnetograms are used for shape-based parameters such as size, directionality, sunspot borders \cite{mcintosh1990classification}, and polarity inversion lines \cite{ji2023systematic}. Although projection effects still prevail in the original magnetogram rasters, deep-learning models can learn from the compressed full-disk images as observed in \cite{pandeyecml2023, pandeyds2023, pandeydsaa2023} and issue the flare forecast for the entire solar disk. Therefore, a full-disk model is appropriate to complement the AR-based counterparts as these models can predict the flares that appear on the near-limb regions of the Sun and add a crucial element to the operational systems.


Deep learning-based approaches have significantly improved results in generic image classification tasks; however, these models are not easily interpretable due to the complex modeling that obscures the rationale behind the model's decision. Understanding the decision-making process is critical for operational flare forecasting systems. Recently, several empirical methods have been developed to explain and interpret the decisions made by deep neural networks. These are post hoc analysis methods (attribution methods) (e.g., \cite{gradcam}), meaning they focus on the analysis of trained models and do not contribute to the model's parameters while training. In this work, we primarily focus on developing a convolutional neural network (CNN) based full-disk model with trainable attention modules that can amplify the relevant features and suppress the misleading ones while predicting $\geq$M1.0-class solar flares as well as evaluating and explaining our model's performance by visualizing the attention maps overlaying on the input magnetograms to understand which regions on the magnetogram were considered relevant for the corresponding decision. To validate and compare our results, we train a baseline model with the same architecture as our attention model, which however, follows the standard CNN pipeline where a global image descriptor for an input image is obtained by flattening the activations of the last convolutional layer. 

By integrating attention modules into the standard CNN pipeline, we attain two significant advantages: enhanced model performance and the ability to gain insight into the decision-making process. This integration not only improves the predictive abilities but also provides an interpretable model that reveals the significant features influencing the model's decisions. The architecture combines the CNN pipeline with trainable attention modules as mentioned in \cite{attn}. Both of our model's architectures are based on the general CNN pipeline; details are described later in Sec.~\ref{sec:model}. The novel contributions of this paper are as follows: (i) We introduce a novel approach of a light-weight attention-based model that improves the predictive performance of traditional CNNs for full-disk solar flare prediction (ii) We utilize the attention maps from the model to understand the model's rationale behind prediction decision and show that the model's decisions are linked to relevant ARs (iii) We show that our models can tackle the prediction of flares appearing on near-limb regions of the Sun. 

The remainder of this paper is organized as follows: In Sec.~\ref{sec:rel}, we outline the various approaches used in solar flare prediction with contemporary work using deep learning. In Sec.~\ref{sec:data}, we explain our data preparation and class-wise distribution for binary prediction mode. In \ref{sec:model} we present a detailed description of our flare prediction model. In Sec.~\ref{sec:expt}, we present our experimental design and evaluations. In Sec.~\ref{sec:dis} we present case-based qualitative interpretations of attention maps, and, lastly, in Sec.~\ref{sec:conc}, we provide our concluding remarks with avenues for future work.

\section{Related Work}\label{sec:rel}
Solar flare prediction currently, to the best of our knowledge, relies on four major strategies: (i) empirical human prediction (e.g., \cite{Crown2012, Devos2014}), which involves manual monitoring and analysis of solar activity using various instruments and techniques, to obtain real-time information about changes in the Sun's magnetic field and surface features, which are often precursors to flare activity; (ii) physics-based numerical simulations (e.g., \cite{Kusano2020, Korss2020}), which involves a detailed understanding of the Sun's magnetic field and the processes that drive flare activity and running simulations models to predict the occurrence of flares;  (iii) statistical prediction (e.g., \cite{Lee2012, Leka2018}), which involves studying the historical behavior of flares to predict their likelihood in the future using statistical analysis and is closely related to (iv) machine learning and deep learning approaches (e.g., \cite{Bobra2015, Nishizuka2018, Nishizuka2021, Huang2018, Li2020, Ji2020, Pandey2021, Pandey2022, Pandey2022f, Whitman2022, Hong2023}), which involves training algorithms with vast amount of historical data and creating data-driven models that detects subtle patterns  associated with flares in solar activity and make predictions. 

The rapid advancements in deep learning techniques have significantly accelerated research in the field of solar flare prediction. A CNN-based flare forecasting model trained with solar AR patches extracted from line-of-sight (LoS) magnetograms within $\pm$30$^{\circ}$ of the central meridian to predict $\geq$C-, $\geq$M-, and $\geq$X-class flares was presented in \cite{Huang2018}. Similarly, \cite{Li2020} use a CNN-based model to issue binary class predictions for both $\geq$C- and $\geq$M-class flares within 24 hours using Space-Weather Helioseismic and Magnetic Imager Active Region Patches (SHARP) data \cite{Bobra2014} extracted from solar magnetograms using AR patches located within $\pm45^{\circ}$ of the central meridian. Both of these models are limited to a small portion of the observable disk in central locations ($\pm30^{\circ}$ and $\pm45^{\circ}$) and thus have limited operational capability.  Moreover, in our previous studies \cite{Pandey2021, Pandey2022}, we presented deep learning-based full-disk flare prediction models. These models were trained using smaller datasets and these proof-of-concept models served as initial investigations into their potential as a supplementary component for operational forecasting systems. More recently, we presented explainable full-disk flare prediction models \cite{pandeyecml2023, pandeyds2023}, utilizing attribution methods to comprehend the models' effectiveness for near-limb flare events. We observed that the deep learning-based full-disk models are capable of identifying relevant areas in a full-disk magnetogram, which eventually translates into the model's prediction. However, these models utilized a post-hoc approach for model explanation, which does not contribute to further improving the model's performance.


In recent years, attention-based models, particularly Vision Transformers (ViTs) \cite{vit}, have emerged as powerful contenders for image classification tasks, achieving competent results on large-scale datasets. ViTs leverage self-attention mechanisms to effectively capture long-range dependencies in images, enabling them to excel in complex visual recognition tasks. While ViTs offer state-of-the-art performance, they often come with a large number (86 to 632 million) of trainable parameters, making them resource-intensive and less practical for scenarios with limited computational resources or small-sized datasets. To address this issue, for our specific use case with a small dataset, we are exploring alternative models that strike a balance between accuracy and efficiency. By incorporating attention blocks into a standard CNN pipeline, we obtain a much lighter model, consisting of $\sim$7.5 million parameters. This approach allows for computationally efficient near-real-time predictions with relatively less resource demand on deployment infrastructure while ensuring competent performance for solar flare prediction compared to our prior work \cite{pandeyds2023, pandeydsaa2023} with customized AlexNet-based \cite{alex} full-disk model, with $\sim$57.25 million parameters and fine-tuned VGG16 \cite{vgg} full-disk model in \cite{pandeyecml2023} with $\sim$134 million parameters.

\section{Data}\label{sec:data}
\vspace{-10pt}
\begin{figure}[tbh!]
\centering
\includegraphics[width=0.85\linewidth ]{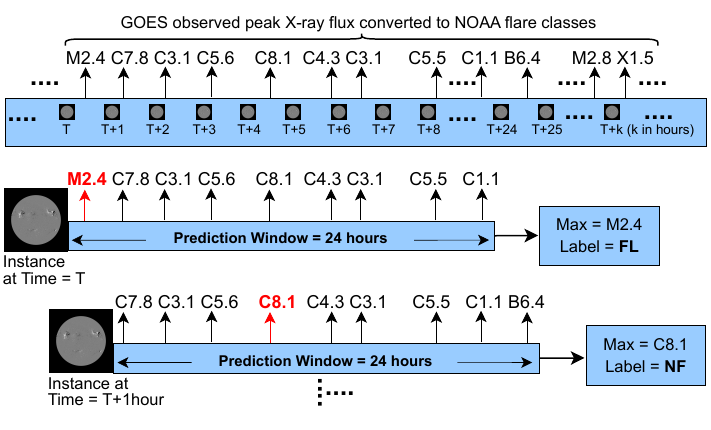}
\caption[]{A visual representation of the data labeling process using hourly observations of full-disk LoS magnetograms and a prediction window of 24 hours considered to label the magnetograms. Here, `FL' and `NF' indicate `Flare' and `No Flare' for binary prediction mode ($\geq$M1.0-class flares).}
\label{fig:timeline}
\vspace{-5pt}
\end{figure}

 We use full-disk line-of-sight (LoS) magnetogram images obtained from the Helioseismic and Magnetic Imager (HMI) \cite{Schou2011} instrument onboard Solar Dynamics Observatory (SDO) \cite{Pesnell2011} publicly available from Helioviewer \cite{Muller2009}. We collected hourly instances of magnetogram images at [00:00, 01:00,...,23:00] each day from December 2010 to December 2018. We labeled the magnetogram images for binary prediction mode ($\geq$M1.0-class flares) based on the peak X-ray flux converted to NOAA flare classes with a prediction window of the next 24 hours. To elaborate, if the maximum of GOES observed peak X-ray flux of a flare is weaker than M1.0, the corresponding magnetogram instances are labeled as ``No Flare'' (NF: $<$M1.0), and larger ones are labeled as ``Flare'' (FL: $\geq$M1.0) as shown in Fig~.\ref{fig:timeline}.
 
\begin{table}[t!]
\setlength{\tabcolsep}{4pt}
\renewcommand{\arraystretch}{1.75}
\caption{The total number of hourly sampled magnetogram images per flare class distributed into four tri-monthly partitions.}
\begin{center}
\begin{tabular}{c c c c c c}
\hline
Binary Class & Partition-1 & Partition-2 & Partition-3 & Partition-4 & Total\\
\hline
NF ($<$M1.0) & 12,454 & 13,855 & 14,308 & 14,032 & 54,649 \\
FL ($\geq$M1.0) & 2,334 & 1,612 & 2,364 & 2,690 & 9,000 \\
\hline
FL:NF & $\sim$1:5 & $\sim$1:9 & $\sim$1:6 & $\sim$1:5 & $\sim$1:6 \\
\hline
\end{tabular}
\label{table:datatable}
\end{center}
\vspace{-17pt}
\end{table}

Our dataset includes a total of 63,649 full-disk LoS magnetogram images, where 54,649 instances belong to the NF-class and 9,000 instances (8,120 instances of M-class and 880 instances of X-class flares) to the FL-class \footnote{The current total count of 63,649 magnetogram observations in our dataset is lower than it should be for the period of December 2010 to December 2018. This is due to the unavailability of some instances from Helioviewer.}. We finally create a non-chronological split of our data into four temporally non-overlapping tri-monthly partitions introduced in \cite{Pandey2021} for our cross-validation experiments. This partitioning of the dataset is created by dividing the data timeline from Dec 2010 to Dec 2018 into four partitions, where Partition-1 contains data from January to March, Partition-2 contains data from April to June, Partition-3 contains data from July to September, and finally, Partition-4 contains data from October to December as shown in Table.~\ref{table:datatable}. Because $\geq$M1.0-class flares are scarce, the data distribution exhibits a significant imbalance, with the highest imbalance occurring in Partition-2 (FL:NF $\sim$1:9). Overall, the imbalance ratio stands at $\sim$1:6 for FL to NF class.
\begin{figure*}[bth!]
  \centering
  \includegraphics[width=0.8\linewidth]{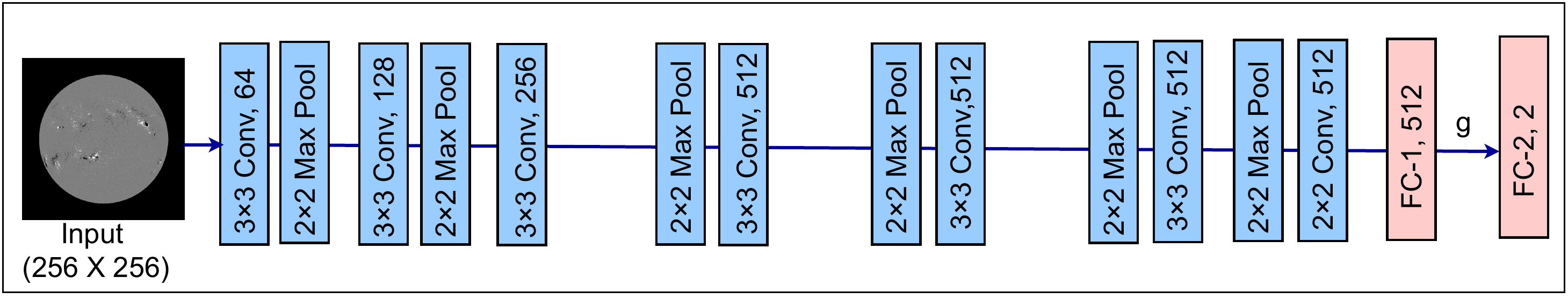}
  \caption{The architecture of our baseline model (M1). Note: Each convolutional layer (except the last one) is followed by a batch normalization layer.}
  \label{fig:arch1}
\end{figure*}

\begin{figure*}[bth!]
  \centering
  \includegraphics[width=0.8\linewidth]{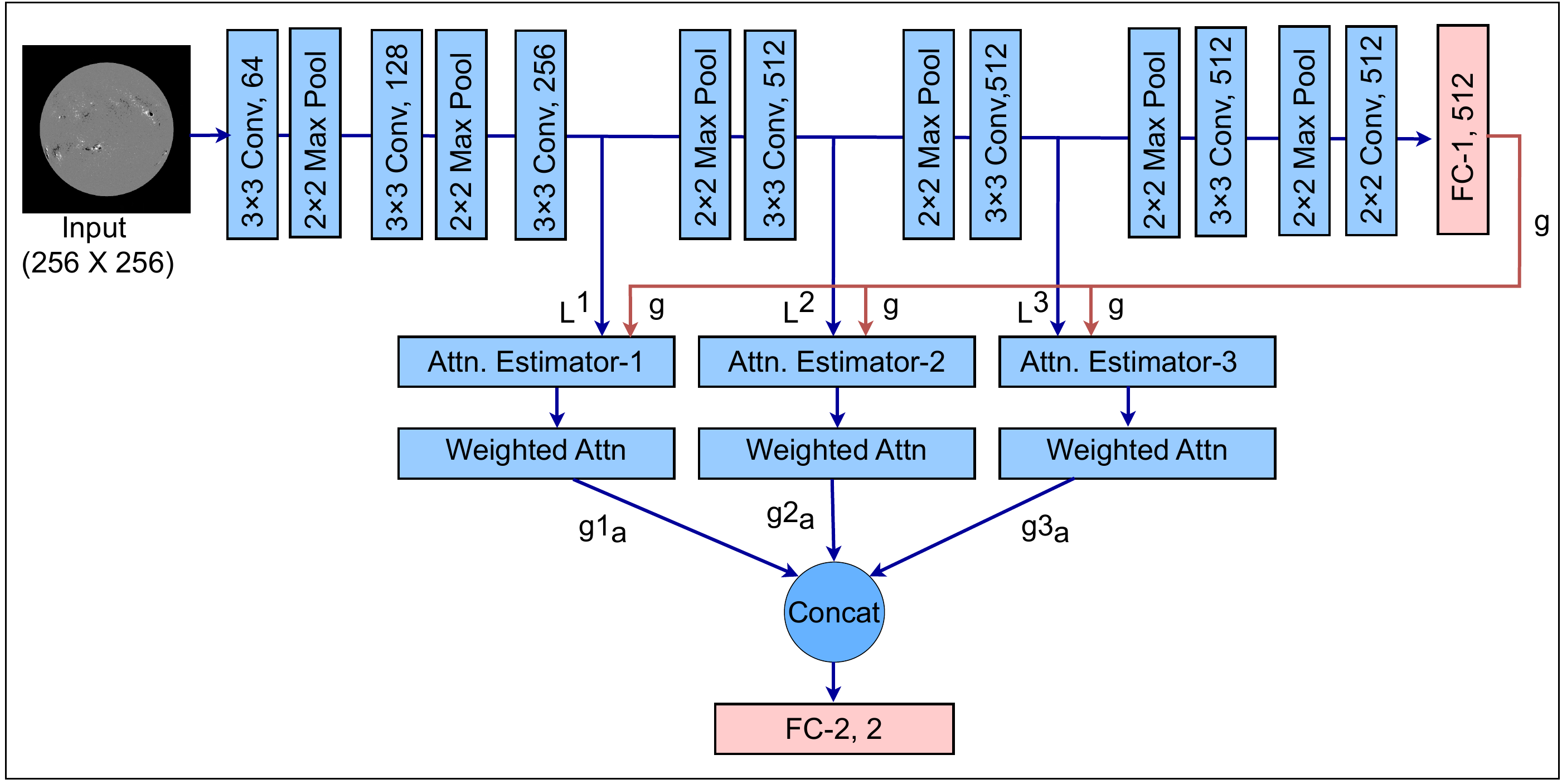}
  \caption{The architecture of our attention-based flare prediction model (M2). The model has three trainable attention modules integrated after the third, fourth, and fifth convolution blocks before the max-pool layer. Note: Each convolutional layer (except the last one) is followed by a batch normalization layer.}
  \label{fig:arch2}
  \vspace{-10pt}
\end{figure*}
\section{Model}\label{sec:model}

In this work, we develop two deep learning models: (i) standard CNN model as a baseline (denoted as \emph{M1}), and (ii) attention-based model (denoted as \emph{M2}) proposed in \cite{attn} to perform and compare in the task of solar flare prediction. The M1 model shown in Fig.~\ref{fig:arch1} follows an intuition of standard CNN architecture where a global image descriptor ($g$) is derived from the input image from the activations of the last convolutional layer and passed through a fully connected layer to obtain class prediction probabilities. On the other hand, the attention-based full-disk model (M2) encourages the filters earlier in the CNN pipeline to learn similar mappings compatible with the one that produces a global image descriptor in the original architecture. Furthermore, it focuses on identifying salient image regions and amplifying their influence while suppressing the irrelevant and potentially spurious information in other regions during training and thus utilizing a trainable attention estimator by integrating it into the standard CNN pipeline. The architecture of our attention-based model is shown in Fig.~\ref{fig:arch2}. The architecture of the attention model proposed in \cite{attn} integrates the trainable attention modules in a modified VGG-16 \cite{vgg} architecture. We use a simpler VGG-like architecture with a reduced number of convolutional layers, which also reduces the number of parameters. Our first convolutional layer accepts a 1-channel input magnetogram image resized to 256$\times$256. Each convolutional layer (except the last one) is followed by a batch normalization layer before max pooling. The final convolutional layer outputs feature maps of size 512$\times$1$\times$1 that squeezed into a fully connected layer (FC-1) with a 512-dimensional vector, which is the global representation ($g$) of the input image.

The M2 model follows the same architecture as in M1, except it has three trainable attention modules integrated after the third, fourth, and fifth convolution blocks before the max-pool layer. The similarity between the architectures is intentional to demonstrate the impact of the attention estimators on model performance. Similarly, integrating attention modules in the middle of the network is also a deliberate design choice. As the early layers in CNN primarily focus on low-level features \cite{Alzubaidi2021}, we position the attention modules further into the pipeline to capture higher-level features. However, there is a tradeoff involved, as pushing attention to the last layers is hindered by significantly reduced spatial resolution in the feature maps. Consequently, placing attention modules in the middle strikes a balance, making it a more suitable and pragmatic approach.  

In the M2 model, outputs from the convolutional blocks (denoted as $L^s$) are passed to the attention estimators. In other words, $L^s$ is a set of feature vectors:
\begin{equation*}
\vspace{-5pt}
    L^s=\{l^s_1, l^s_2,..., l^s_n\}
\end{equation*}
extracted at a given convolutional layer to serve as input to the $s_{th}$ attention estimator, and $l^s_i$ is the vector of output activations at $i^{th}$ of total $n$ spatial locations in the layer. $g$ represents a global feature vector obtained by flattening the feature maps at the first fully connected layer, located at the end of the convolution blocks (referred to as FC-1 in Fig.\ref{fig:arch2}).

The attention mechanism aims to compute a compatibility score, denoted as $C(L^s, g)$, utilizing the local features ($L^s$) and global feature representations ($g$), and replaces the final feature vector with a set of attention-weighted local features. As the compatibility scores $C$ and $L^s$ are required to have the same dimension, the dimension matching is performed by a linear mapping of vectors $l^s_i$ to the dimension of $g$. Then, the compatibility function $C(L^s, g)=\{c^s_1, c^s_2,..., c^s_n\} $ is a set for each vector $l^s_i$, which is computed as an addition operation (additive attention) as follows:
\begin{equation*}
c^s_1 = (l^s_i, g), \text{ for } i \in \{1, 2, ..., n\}.    
\end{equation*}

The computed compatibility scores are then normalized using a softmax operation and represented as:
\begin{equation*}
A^s=\{a^s_1, a^s_2,..., a^s_n\}.    
\end{equation*}
The normalized compatibility scores are then used to compute an element-wise weighted average, which results in a vector:
\begin{equation*}
    g^s_a=\sum_{i=1}^{n}  a^s_i.l^s_i
\end{equation*}
for each attention layer, $s$. Finally, the individual $g^s_a$ vectors of size 512 are concatenated to get a new attention-based global representation to perform the binary classification in the (second) fully connected layer (FC-2). This approach allows the activations from earlier layers to influence and contribute to the final global feature representation, thereby enhancing the model's ability to capture relevant spatial information.
\section{Experimental Evaluation}\label{sec:expt}
\subsection{Experimental Settings}
We trained both of our models (M1 and M2) with stochastic gradient descent (SGD) as an optimizer and cross-entropy as the objective function. Both models are initialized using Kaiming initialization from a uniform distribution \cite{kai}, and then we use a dynamic learning rate (initialized at 0.001 and reduced by half every 3 epochs) to further train the model to 40 epochs with a batch size of 128. We regularized our models with a weight decay parameter tuned at 0.5 to prevent overfitting. As mentioned earlier in Sec.~\ref{sec:data}, we are dealing with an imbalanced dataset. Therefore, we address the class imbalance problem through data augmentation and oversampling exclusively for the training set while maintaining the imbalanced nature of the test set for realistic evaluation. Firstly, we use three augmentation techniques: vertical flipping, horizontal flipping, and +5$^{\circ}$ to -5$^{\circ}$ rotations on minority class (FL-class) which decreases the imbalance from 1:6 to approximately 2:3. Finally, we randomly oversampled the minority class to match the instances of NF-class resulting in a balanced dataset. We prefer augmentation and oversampling over undersampling as the flare prediction models trained with undersampled data are shown to lead to inferior performance \cite{Ahmadzadeh2021} (usually transpiring as one-sided predictions). We employed a 4-fold cross-validation schema for validating our models, using the tri-monthly partitions (described in Sec.~\ref{sec:data}), where we applied three partitions for training the model and one for testing. 

We evaluate the performance of our models using two widely-used forecast skills scores: True Skill Statistics (TSS, in Eq.~\ref{eq:TSS}) and Heidke Skill Score (HSS, in Eq.~\ref{eq:HSS}), derived from the elements of confusion matrix: True Positives (TP), True Negatives (TN), False Positives (FP), and False Negatives (FN). In the context of this paper, the ``FL-class''  is considered as the positive outcome, while the ``NF-class'' is negative.

\begin{equation}\label{eq:TSS}
    TSS = \frac{TP}{TP+FN} - \frac{FP}{FP+TN} 
\end{equation}

\begin{equation}\label{eq:HSS}
    HSS = 2\times \frac{TP \times TN - FN \times FP}{((P \times (FN + TN) + (TP + FP) \times N))}
\end{equation}
,where N = TN + FP and  P = TP + FN. TSS and HSS values range from -1 to 1, where 1 indicates all correct predictions, -1 represents all incorrect predictions, and 0 represents no skill. In contrast to TSS, HSS is an imbalance-aware metric, and it is common practice to use HSS for the solar flare prediction models due to the high class-imbalance ratio present in the datasets. For a balanced test dataset, these metrics are equivalent \cite{Ahmadzadeh2021}. Lastly, we report the subclass and overall recall for flaring instances (M- and X-class), which is calculated as ($\frac{TP}{TP+FN}$), to demonstrate the prediction sensitivity.

\subsection{Evaluation}
\begin{table}[tbh!]
\setlength{\tabcolsep}{6pt}
\renewcommand{\arraystretch}{1.5}
\vspace{-10pt}
\caption{Average performance of our models in terms of two skill scores (TSS and HSS) evaluated on the test set for the 4-fold cross-validation experiment.}
\begin{center}
\begin{tabular}{c c c}
\hline
Models  & TSS & HSS\\
\hline
M1 & 0.35$\pm$0.13 & 0.30$\pm$0.09\\
Pandey et al. \cite{pandeyecml2023} & $\sim$0.51 & $\sim$0.35\\
Pandey et al. \cite{pandeyds2023} & 0.51$\pm$0.05 & 0.38$\pm$0.08\\
M2 & \textbf{0.54$\pm$0.03}  & \textbf{0.37$\pm$0.07}\\
\hline
\end{tabular}
\label{table:scoretable}
\end{center}
\vspace{-10pt}
\end{table}
We perform a 4-fold cross-validation using the tri-monthly separated dataset for evaluating our models. With the baseline model (M1) we obtain on an average TSS$\sim$0.35$\pm$0.13 and HSS$\sim$0.30$\pm$0.09. The M1 model following the standard CNN pipeline has fluctuations across folds and hence a high margin of error on skill scores is represented by the standard deviation.  Model M2 improves over the performance of model M1 by $\sim$20\% and $\sim$7\% in terms of TSS and HSS respectively. Furthermore, it improves on the performance of \cite{pandeyds2023, pandeyecml2023} by $\sim$3\% in terms of TSS and shows comparable results in terms of HSS and is more robust as indicated by the deviations across the folds as shown in Table \ref{table:scoretable}. Moreover, the performance of model M2 becomes even more noteworthy when considering its parameter efficiency. With only $\sim$7.5 million parameters, it outperforms \cite{pandeyds2023} an AlexNet-based model and \cite{pandeyecml2023} a VGG16-based model with a much higher parameter count of $\sim$57.25  and $\sim$134 million respectively, showcasing the effectiveness of attention mechanisms in achieving superior results while maintaining a significantly leaner model architecture. This highlights the potential of this approach to provide both performance gains and resource optimization. The findings of this study emphasize the significance of optimizing attention configurations to enhance model performance, taking into account both parameter complexities and the strategic combination of attention patterns for effective pattern recognition. 

\begin{table}[tbh!]
\setlength{\tabcolsep}{3pt}
\renewcommand{\arraystretch}{1.5}
\caption{Counts of correctly (TP) and incorrectly (FN) classified X- and M-class flares in central ($|longitude|$$\leq\pm70^{\circ}$) and near-limb locations. The recall across different location groups is also presented. Counts are aggregated across folds.}
\begin{center}
 \begin{tabular}{r r c c c c c c}
\hline
 & &
\multicolumn{3}{c}{Within $\pm$70$^{\circ}$} 
&                                            
\multicolumn{3}{c}{Beyond $\pm$70$^{\circ}$}\\
Models & Flare-Class & TP  & FN  & Recall  & TP   &FN & Recall \\
\hline
& X-Class  &  467  & 201  & 0.70 & 100 & 112 & 0.47\\

M1 & M-Class &  3153 & 2677  & 0.54 & 878   & 1412 & 0.38\\

& Total (X\&M) & 3620 & 2878 &0.62 & 978 & 1524 & 0.43\\ 
\hline
&X-Class  &  636  & 32  & \textbf{0.95} & 164 & 48 & \textbf{0.77}\\

M2 &M-Class &  4850 & 980  & \textbf{0.83} & 1161   &1129 & \textbf{0.51}\\

&Total (X\&M) & 5486 & 1012 & \textbf{0.89} & 1325 & 1177 & \textbf{0.64}\\ 
\hline
\end{tabular}
\end{center}
\label{table:comp}
\vspace{-5pt}
\end{table}

\begin{figure}[tbh!]
\centering
\includegraphics[width=0.98\linewidth ]{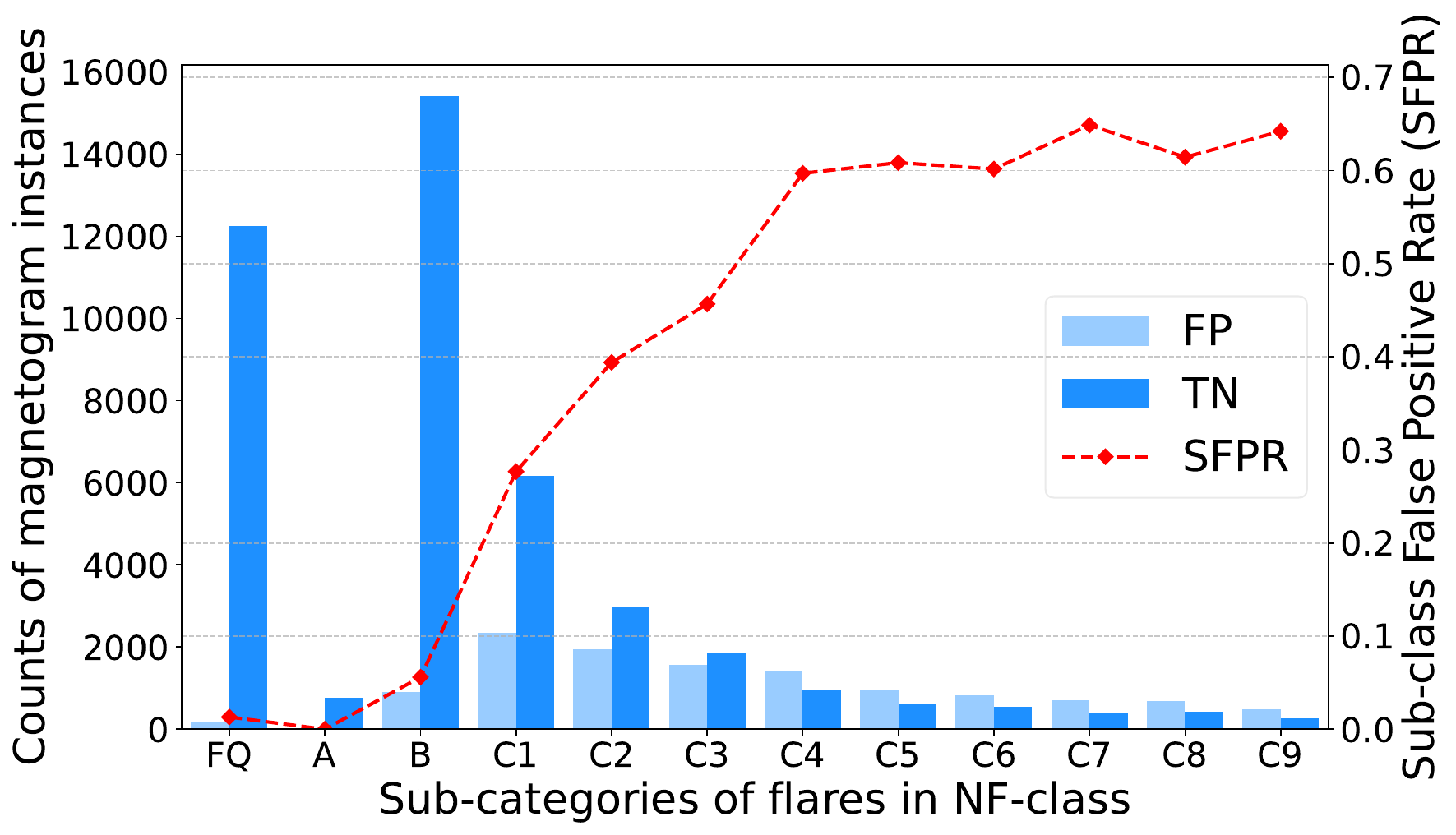}
\caption[]{A bar-line plot showing the true negatives (TN), false positives (FP), and false positive rate for sub-classes in NF-class (SFPR) obtained from model M2. The results are aggregated from validation sets of 4-fold experiments. }
\label{fig:falsepositives}
\vspace{-10pt}
\end{figure}

In addition, we evaluate our results for correctly predicted and missed flare counts for class-specific flares (X-class and M-class) in central locations (within $\pm$70$^{\circ}$) and near-limb locations (beyond $\pm$70$^{\circ}$) of the Sun as shown in Table \ref{table:comp}. We observe that the attention-based model (M2) shows significantly better results compared to the baseline (M1). The M2 model made correct predictions for $\sim$95\% of the X-class flares and $\sim$83\% of the M-class flares in central locations. Similarly, it shows a compelling performance for flares appearing on near-limb locations of the Sun, where $\sim$77\% of the X-class and $\sim$51\%  of the M-class flares are predicted correctly. This is important because, to our knowledge, the prediction of near-limb flares is often overlooked, although vital for predicting Earth-impacting space weather events. More false negatives in M-class are expected because of the model's inability to distinguish bordering class (C4+ to C9.9) flares from $\geq$M1.0-class flares as shown in Fig.~\ref{fig:falsepositives}. We observed an upward trend in the false positive rate for sub-classes (SFPR) within C-class flares when compared to other sub-classes, such as Flare-Quiet (FQ), A-class, and B-class flares. More specifically we note that the count of false positives (FP) surpasses that of true negatives (TN) for flare classes ranging from $\geq$C4 to $\leq$C9. 
The prevalence of FP in $\geq$C4-class flares suggests a need for improved predictive capabilities between border classes. 

Overall, we observed that our model predicted $\sim$89\% of the flares in central locations and $\sim$64\% of the flares in near-limb locations. Furthermore, class-wise analysis shows that $\sim$91\% and $\sim$74\% of the X-class and M-class flares, respectively, are predicted correctly by our models. To reproduce this work, the source code is available in our open-source repository \cite{sourcecode}.

\begin{figure*}[tbh!]
  \centering
  \includegraphics[width=0.78\linewidth]{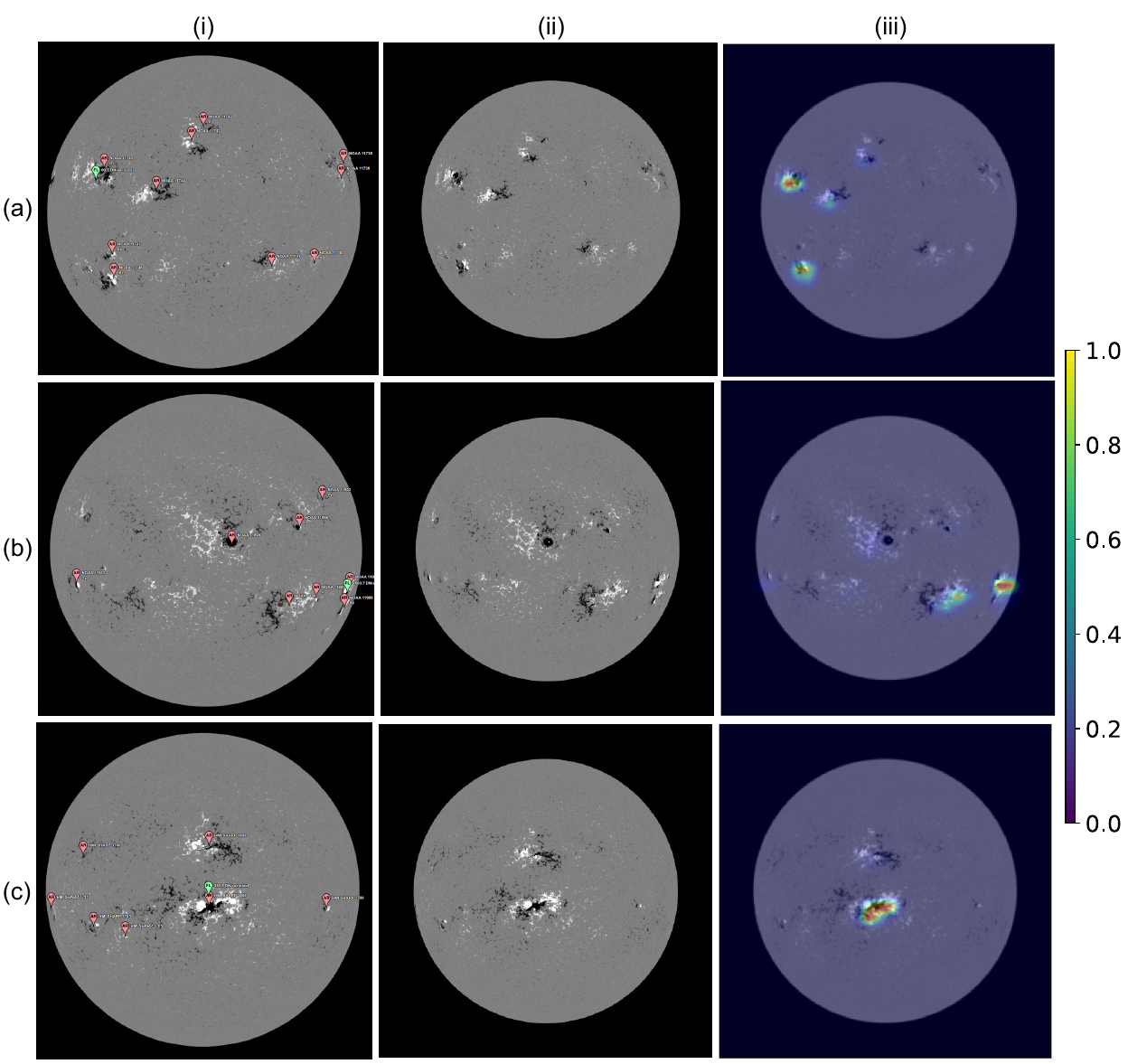}
  \caption{A figure-grid of case-based visual interpretation for three different instances using attention maps, each represented in a separate row indexed as (a), (b), and (c). Row (a) shows correctly predicted near-limb (East) FL-class, (b) shows correctly predicted near-limb (West) FL-class, and (c) shows incorrectly predicted NF-class instances. In each row: column (i) displays an annotated full-disk magnetogram image at the onset of the flare, with green flags indicating the flare's location and red flags representing all ARs present in the magnetogram. Column (ii) shows an actual magnetogram image used in our dataset to train the model. Finally, column (iii) depicts a visualization created by overlaying the attention maps obtained from Attention Estimator-2. The color bar shows the scale of normalized attention map values ranging from 0-1, where a higher value suggests important features for a corresponding prediction.}
  \label{fig:attention}
  \vspace{-10pt}
\end{figure*}

\section{Discussion}\label{sec:dis}
In this section, we visualize the attention maps learned by the M2 model to qualitatively analyze and understand regions in input magnetogram images that are considered relevant. We applied three attention layers in our model M2, where attention maps $(L1, L2, L3)$ has a spatial dimension $(\frac{1}{4}, \frac{1}{8}, \frac{1}{16})th$ of the input size respectively. To visualize the relevant features learned by the models using attention layers, we upscale these maps to the size of the magnetogram image using bilinear interpolation and overlay the maps on top of the original image. We present the attention maps from the Attention Estimator-2 because the first attention layer focuses on lower-level features, which are scattered and do not provide a globally detailed explanation. On the other hand, the Attention Estimator-3 focuses on higher-level features, and due to the high reduction in spatial dimension ($\frac{1}{16}$ of the original input), upscaling through interpolation results in a spatial resolution that is insufficient for generating interpretable activation maps. 

As the primary focus of this study is to understand the capability of full-disk models on the near-limb flares, we showcase a near-limb (East) X3.2-class flare observed on 2013-05-14T00:00:00 UTC. Note that East and West are reversed in solar coordinates. The location of the flare is shown by a green flag in Fig.~\ref{fig:attention} (a)(i), along with the ARs (red flags). For this case-based qualitative analysis, we use an input image at 2013-05-13T06:00:00 UTC ($\sim$18 hours prior to the flare event), shown in Fig.~\ref{fig:attention} (a)(ii) and in Fig.~\ref{fig:attention} (a)(iii), we show the overlaid attention map, which pinpoints important regions in the input image where specific ARs are activated as relevant features, suppressing a large section of the full-disk magnetogram disk although there are 10 ARs (red flags). More specifically, the model focuses on the same AR that is responsible for initiating a flare 18 hours later. Similarly, we analyze another case of correctly predicted near-limb (West) X1.0-class flare observed on 2013-11-19T10:14:00 UTC shown in Fig.~\ref{fig:attention} (b)(i). For this, we used an input image at 2013-11-18T17:00:00 UTC ($\sim$17 hours prior to the flare event) shown in Fig.~\ref{fig:attention} (b)(ii). We again observed that the model focuses on the relevant AR even though other, relatively large ARs are present in the magnetogram image as shown in Fig.~\ref{fig:attention} (b)(iii). 

Furthermore, we provide an example to analyze a case of false positives as well. For this, we use an example of a C7.9 flare observed on 2014-02-03T00:12:43 UTC shown in Fig~.\ref{fig:attention} (c)(i), and to explain the result, we used an input magnetogram instance at 2014-02-02T23:00:00 UTC ($\sim$14 hours prior to the event) shown in Fig~.\ref{fig:attention} (c)(ii). For the given time, there are 7 ARs indicated by the red flags, however, on interpreting this prediction with attention maps shown in Fig~.\ref{fig:attention} (c)(iii), we observed that the model considers only one region as a relevant feature for the corresponding prediction, which is indeed the location of the C7.9 flare. This incorrect prediction can be attributed to interference caused by bordering C-class flares as shown earlier in Fig.~\ref{fig:falsepositives}, where we noted that among the 25,150 C-class flares observed, $\sim$43\% (10,935) resulted in incorrect predictions, constituting $\sim$91\% of the total false positives.
\section{Conclusion and Future Work}\label{sec:conc}
In this work, we presented an attention-based full-disk model to predict $\geq$M1.0-class flares in binary mode and compared the performance with standard CNN-based models. We observed that the trainable attention modules play a crucial role in directing the model to focus on pertinent features associated with ARs while suppressing irrelevant features in a magnetogram during training, resulting in an enhancement of model performance. Furthermore, we demonstrated, both quantitatively through recall scores and qualitatively by overlaying attention maps on input magnetogram images, that our model effectively identifies and localizes relevant AR locations, which are more likely to initiate a flare. This prediction capability extends to near-limb regions, making it crucial for operational systems. As an extension, we plan to include the temporal aspects in our dataset and create a spatiotemporal model to capture the evolution of solar activity leading to solar flares. Furthermore, we plan to extend this work by developing an automated way of analyzing the interpretation results to identify the main causes of incorrect predictions. 

\section*{Acknowledgments}
This project is supported in part under two NSF awards \#2104004 and \#1931555, jointly by the Office of Advanced Cyberinfrastructure within the Directorate for Computer and Information Science and Engineering, the Division of Astronomical Sciences within the Directorate for Mathematical and Physical Sciences, and the Solar Terrestrial Physics Program and the Division of Integrative and Collaborative Education and Research within the Directorate for Geosciences. This work is also partially supported by the National Aeronautics and Space Administration (NASA) grant award \#80NSSC22K0272. The data used in this study is a courtesy of NASA/SDO and the AIA, EVE, and HMI science teams, and the NOAA National Geophysical Data Center (NGDC).

\bibliographystyle{IEEEtran}
\bibliography{ref}
\end{document}